\documentclass[lettersize,journal, twoside]{IEEEtran}
\usepackage{amsmath,amsfonts}
\usepackage{algorithmic}
\usepackage{algorithm}
\usepackage{array}
\usepackage[caption=false,font=normalsize,labelfont=sf,textfont=sf]{subfig}
\usepackage{textcomp}
\usepackage{stfloats}
\usepackage{url}
\usepackage{verbatim}
\usepackage{graphicx}
\usepackage{hyperref}
\usepackage{cite}
\usepackage{xspace}
\usepackage{pdflscape}
\usepackage{wrapfig}
\usepackage{xcolor}   % for coloring
\usepackage{booktabs}
\newif\ifrevision

\revisionfalse

\colorlet{revision}{black}
\ifrevision \colorlet{revision}{blue}\fi

  {\begin{figure}[#1]%
   \ifrevision
     \color{revision}%
     \captionsetup{labelfont={color=revision}, textfont={color=revision}}%
   \fi}
  {\end{figure}}

\newenvironment{revisionfigure*}[1][]%
{\begin{figure*}[#1]%
\ifrevision
 \color{revision}%
 \captionsetup{labelfont={color=revision}, textfont={color=revision}}%
\fi}
{\end{figure*}}

\newenvironment{revisiontable}[1][]%
  {\begin{table}[#1]%
   \ifrevision
     \color{revision}%
     \captionsetup{labelfont={color=revision}, textfont={color=revision}}%
   \fi}
  {\end{table}}

\newenvironment{revisiontabular}[1]%
  {\begingroup
   \ifrevision \color{revision}\fi
   \begin{tabular}{#1}}
  {\end{tabular}%
   \endgroup}

\newcommand{\method}{SAMChat\xspace}
\newcommand{\dataset}{SAMData\xspace}

\begin{document}

\title{\textcolor{revision}{SAMChat: Introducing Chain of Thought Reasoning and GRPO to a Multimodal Small Language Model for Small Scale Remote Sensing}}

\author{Aybora Köksal,~\IEEEmembership{Graduate Student Member,~IEEE,}
A. Aydın Alatan,~\IEEEmembership{Senior Member,~IEEE}
        % <-this % stops a space
\thanks{Aybora Köksal and A. Aydın Alatan are with Center for the Image Analysis (OGAM) and Department of Electrical and Electronics Engineering of Middle East Technical University (METU), Ankara, Türkiye (e-mail: aybora@metu.edu.tr, alatan@metu.edu.tr}% <-this % stops a space
%\thanks{Manuscript received April 19, 2021; revised August 16, 2021.}}
%}
\thanks{Manuscript submitted to JSTARS on April 2, 2025; revised September 19, 2025.}}
\markboth{IEEE JSTARS Special Issue on Foundation and Large Vision Models for Remote Sensing, September~2025}{Köksal \MakeLowercase{\textit{et al.}}: SAMChat: Introducing C\MakeLowercase{o}T Reasoning and GRPO to an MSLM for Small Scale Remote Sensing}
%\markboth{Arxiv Preprint, 2025}{Köksal \MakeLowercase{\textit{et al.}}: SAMChat: Introducing C\MakeLowercase{o}T Reasoning and GRPO to an MSLM for Small Scale Remote Sensing}


\maketitle

\begin{abstract}
Remarkable capabilities in understanding and generating text-image content have been demonstrated by recent advancements in multimodal large language models (MLLMs). However, their effectiveness in specialized domains–particularly those requiring resource-efficient and domain-specific adaptations–has remained limited. In this work, a lightweight multimodal language model termed \method is introduced, specifically adapted to analyze remote sensing imagery in secluded areas, including challenging missile launch sites. A new dataset, \dataset, was compiled by verifying hundreds of aerial images through expert review, and subtle military installations were highlighted via detailed captions. Supervised fine-tuning on a 2B-parameter open-source MLLM with chain-of-thought (CoT) reasoning annotations was performed, enabling more accurate and interpretable explanations. Additionally, Group Relative Policy Optimization (GRPO) was leveraged to enhance the model’s ability to detect critical domain-specific cues--such as defensive layouts and key military structures--while minimizing false positives on civilian scenes. Through empirical evaluations, it has been shown that \method significantly outperforms both larger, general-purpose multimodal models and existing remote sensing–adapted approaches on open-ended captioning and classification metrics. Over 80\% recall and 98\% precision were achieved on the newly proposed \dataset benchmark, underscoring the potency of targeted fine-tuning and reinforcement learning in specialized real-world applications. Code, dataset and models are available at \url{https://github.com/aybora/SAMChat}.
\end{abstract}

\section{Introduction}
\label{sec:intro}

Recent AI advancements have revolutionized natural language processing and computer vision through Large Language Models (LLMs) \cite{radford2018improving, Brown2020, achiam2023gpt, touvron2023llama, bai2023qwentechnicalreport, 2023internlm} and Multimodal Large Language Models (MLLMs) \cite{gpt4v, hurst2024gpt, liu2023llava, wang2024qwen2, chen2024internvl}. These models excel at text generation and visual processing. During the past year, GPT-4V \cite{gpt4v} has shown remarkable image comprehension and reasoning, while open-source models like Qwen2-VL \cite{wang2024qwen2} and InternVL2 \cite{chen2024internvl} have rivaled or surpassed proprietary alternatives.

Despite advancements, large-scale models remain generalists, limiting their effectiveness for specialized tasks. Their high computational demands also hinder deployment in resource-constrained settings. This has driven interest in more efficient alternatives--Small Language Models (SLMs) \cite{liu2024mobilellmoptimizingsubbillionparameter, bai2023qwentechnicalreport, 2023internlm, gemmateam2024gemmaopenmodelsbased, gemmateam2024gemma2improvingopen} and Multimodal Small Language Models (MSLMs) \cite{beyer2024paligemma, steiner2024paligemma, wang2024qwen2, chen2024internvl, deitke2024molmo}. These compact models achieve strong performance via domain adaptation while reducing computational overhead, making them ideal for applications requiring privacy, low latency, and energy efficiency.

\begin{figure}[t!]
    \centering
    \includegraphics[width=\linewidth]{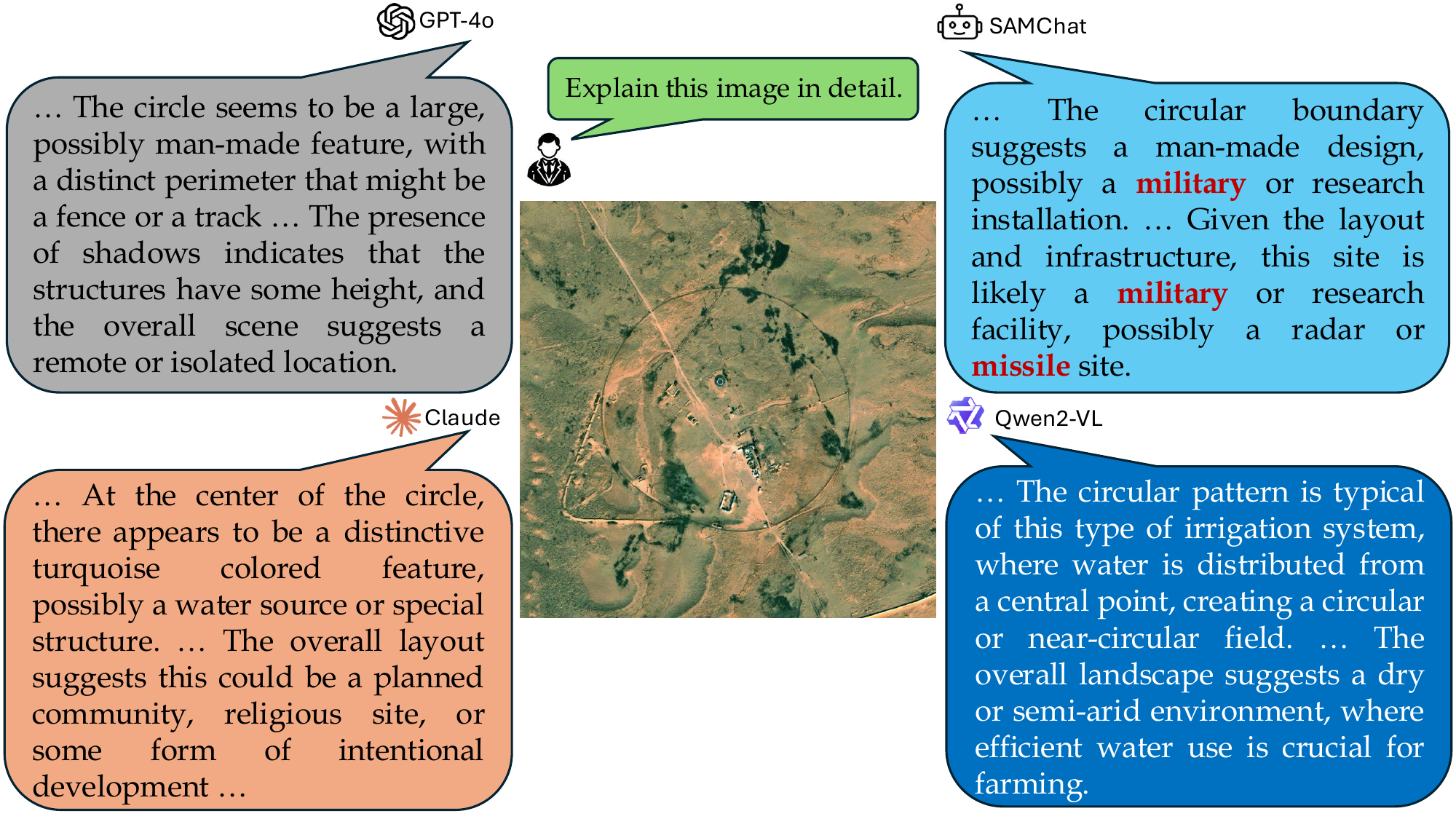}
    \caption{Proposed \method, a CoT and GRPO powered language model for remote sensing, provides significant improvements on secluded area captioning, specifically, on military areas and missile launch sites.}
    \label{fig:teaser}
    \vspace{-.5cm}
\end{figure}

A key question arises: \textit{Can a smaller multimodal model, fine-tuned on targeted data, outperform a much larger model used in zero-shot settings?} Fine-tuning enables domain adaptation, potentially giving specialized models an advantage. Recent NLP studies support this notion, showing that smaller, fine-tuned models sometimes surpass larger ones used without fine-tuning \cite{bucher2024fine}. OpenAI reported that fine-tuned GPT-3.5 Turbo could "match or even outperform" base GPT-4 on specialized tasks--suggesting strategic specialization can sometimes be more effective than model scale alone \cite{gpt35turbo}.

For complex tasks, chain-of-thought reasoning improves accuracy \cite{nye2021show, wei2022chain}. To mitigate error propagation, reinforcement learning and process supervision enhance self-correction and reliability \cite{lightman2023let, uesato2022solving, wang2023math}. Advances in both closed-source models such as OpenAI’s o-series \cite{o1preview, o3mini} and open-source efforts such as DeepSeek-R1 \cite{guo2025deepseek}, aided by Group Relative Policy Optimization (GRPO) \cite{shao2024deepseekmath}, show that even smaller models can be adapted for specialized multimodal reasoning \cite{openr1, openr1multimodal, openr1v}.

With MLLMs excelling in visual tasks, interest is growing in their use for remote sensing (RS) imagery. Several RS-specific MLLMs \cite{hu2307remote, kuckreja2024geochat, zhang2024earthgpt, bazi2024rs, muhtar2024lhrs} tackle tasks including visual question answering, detection, and grounding. However, they mainly perform well on prompt-guided tasks but struggle with open-ended questions. Most are also trained on easily captioned residential data, limiting their effectiveness in remote areas.

%\IEEEpubidadjcol

To address these challenges, this study evaluates MLLM performance on secluded areas, focusing on missile launch sites. An open MLLM captioned expert-verified aerial images to create the \dataset-300 benchmark. A smaller MSLM was fine-tuned into \method-Base, with captions converted to Chain-of-Thought answers for improved reasoning. Using GRPO, the final model, \method-R1, generates concise explanations for aerial imagery.

Based on the experiments, leading generalist MLLMs and RS-specific models struggle with secluded area aerial imagery. However, the proposed \method-R1 demonstrates robustness on the \dataset test set with over 80\% Recall and 98\% Precision. Using DeepSeek-R1's approach with SFT on reasoning data and GRPO, \method-R1 effectively explains any aerial imagery while remaining lightweight for edge devices in both military and residential settings. At 2B parameters, it is the most compact edge-deployable RS-specific language model.

The contributions of the paper are as follows: 

\begin{itemize}
    \item We introduce \dataset, a challenging dataset with captions on expert-verified satellite imagery of secluded areas, specifically missile launch sites. These sites are highly diverse and consist of hard-to-interpret regions due to their complex nature.
    \item We propose a GRPO-based keyword reward function, which, to our knowledge, is the first RL-based training approach for an RS-specific multimodal language model.
    \item We introduce the \method model family, a set of task-specific 2B parameter MSLMs designed to answer open-ended questions about aerial imagery.
    \item Through rigorous experiments, we demonstrate that our model outperforms existing generalist and RS-specific MLLMs, even those with significantly larger parameter sizes, on open-ended questions--measured using keyword-based precision and recall scores.
\end{itemize}

\section{Related Work}
\label{sec:related}

\begin{table}\footnotesize
\centering
\caption{A comparative study of our approach with prior work. Open: Open sourced model, RS Adapted: Adapted to remote sensing problems with domain specific data, RL Reasoning: Trained as reasoning model with an RL based method \label{table:comparative_summary}.}
\begin{tabular}{c|c|c|c|c} \toprule
     Method & Year & Open & RS Adapted & RL Reasoning \\ \midrule
     RSGPT \cite{bazi2024rs} & 2023 & \checkmark & \checkmark & $\times$ \\
     GPT-4o \cite{hurst2024gpt} & 2024 & $\times$ & $\times$ & $\times$ \\ 
     Qwen2-VL \cite{wang2024qwen2} &  2024 & \checkmark & $\times$ & $\times$  \\
     EarthGPT \cite{zhang2024earthgpt} & 2024 & \checkmark & \checkmark & $\times$ \\
     GeoChat \cite{kuckreja2024geochat} & 2024 & \checkmark & \checkmark & $\times$ \\ 
     LHRSBot \cite{muhtar2024lhrs} & 2024 & \checkmark & \checkmark & $\times$ \\ 
     o1-o3 \cite{o1preview} & 2025 & $\times$ & $\times$ & \checkmark \\
     \midrule
     \method  & \textbf{Ours} & \checkmark & \checkmark & \checkmark \\ \bottomrule
\end{tabular}
\vspace{-.5cm}
\end{table}

\textbf{Large Language Models.} The introduction of Transformers \cite{Vaswani2017} revolutionized LLMs, enabling bidirectional context modeling in BERT \cite{Devlin2019} and autoregressive pretraining in GPT \cite{radford2018improving}. These advances led to state-of-the-art models including GPT-3 \cite{Brown2020} with human-like text generation capabilities. The LLaMA family \cite{touvron2023llama} proved that smaller, efficiently trained models could outperform larger ones like GPT-3. Open-source alternatives such as Qwen \cite{bai2023qwentechnicalreport}, InternLM \cite{2023internlm}, and Google's Gemma \cite{gemmateam2024gemmaopenmodelsbased} have since expanded access to powerful LLMs.

Early advancements in Vision-Language Models (VLMs) such as CLIP \cite{radford2021learning} used contrastive learning to align images and text, improving zero-shot capabilities. Integration of VLMs with LLMs created Multimodal Large Language Models (MLLMs), exemplified by GPT-4V \cite{gpt4v}. In open-source, the LLaVA family \cite{liu2023llava} combined CLIP with LLaMA architectures, with LLaVA-NeXT \cite{liu2024llavanext} enhancing visual perception through high-resolution inputs and improved OCR. Other contributions include Qwen-VL \cite{bai2023qwen}, InternVL \cite{chen2024internvl}, and Google's PaliGemma \cite{beyer2024paligemma}, while Molmo \cite{deitke2024molmo} establishes transparency by releasing complete training data and evaluation frameworks.

Multimodal Small Language Models (MSLMs) are compact multimodal models ($<$7B parameters) optimized for efficiency and edge deployment. Unlike resource-intensive MLLMs such as GPT-4o \cite{hurst2024gpt} and InternVL2-72B \cite{chen2024internvl}, MSLMs balance performance with accessibility \cite{beyer2024paligemma, wang2024qwen2}. Recent examples include PaliGemma \cite{beyer2024paligemma}, PaliGemma 2 \cite{steiner2024paligemma}, and Qwen2-VL-2B \cite{wang2024qwen2}, with models as small as 2-3B parameters. InternVL2.5 \cite{chen2024internvl} introduced 1B and 2B variants, while MolmoE-1B \cite{deitke2024molmo} uses a Mixture of Experts approach with just 1.5B active parameters.

While MSLMs were initially designed as generalist models, they excel in task-specific roles. Recent studies \cite{bucher2024fine, gpt35turbo} suggest that small language models can outperform larger ones when specialized for specific tasks, making them advantageous for efficiency and adaptability.

\noindent \textbf{CoT Prompting and Reasoning Models.} The concept of reasoning in LLMs stems from Chain-of-Thought (CoT) prompting. Nye et al. \cite{nye2021show} introduced the Scratchpad Approach for multi-step computations. Wei et al. \cite{wei2022chain} formalized CoT, showing significant improvements in various reasoning tasks for large models. Wang et al. \cite{wang2022self} proposed Self-Consistency, aggregating multiple reasoning paths. Gao et al. \cite{gao2023pal} developed Program-aided Language Models using Python for reasoning, while Yao et al. \cite{yao2023tree} created Tree of Thoughts, structuring reasoning as a tree for evaluation and backtracking.

Zhang et al. \cite{zhang2023multimodal} introduce Multimodal Chain-of-Thought, integrating text and images for complex reasoning. Hu et al. \cite{hu2025visual} propose Visual Sketchpad for LLMs to generate visual representations, while Li et al. \cite{li2025imagine} develop Multimodal Visualization-of-Thought (MVoT) using image-based visualizations. These advances demonstrate the increasing significance of multimodal reasoning capabilities in LLMs.

CoT reasoning is the most effective in large models, with smaller ones struggling on complex tasks \cite{wei2022chain}. To mitigate error propagation \cite{lightman2023let}, researchers have developed several approaches: Process Supervision using step-by-step human feedback \cite{lightman2023let}, process-based supervision \cite{uesato2022solving}, Math-Shepherd for reasoning enhancement \cite{wang2023math}, SCoRe for self-correction \cite{kumar2024training}, TS-LLM's tree-search framework \cite{feng2023alphazero}, and AlphaGeometry for theorem proving \cite{trinh2024solving}.

While math-focused reasoning models have advanced, OpenAI's o1 \cite{o1preview} and o3-mini \cite{o3mini} series have set new benchmarks using reinforcement learning. DeepSeek's open-source DeepSeek-R1 \cite{guo2025deepseek} improves reasoning without supervised fine-tuning. Its DeepSeek-R1-Zero variant uses large-scale RL for self-evolving reasoning but faces language mixing issues. DeepSeek-R1 addresses these with cold-start data and multi-stage training, matching OpenAI-o1-1217's performance. Using Group Relative Policy Optimization (GRPO) \cite{shao2024deepseekmath}, it enhances mathematical reasoning while maintaining efficiency. Through distillation, DeepSeek-R1's capabilities transfer to smaller models (1.5B--70B parameters), outperforming previous models.

Although these reasoning models outperform standard language models, they remain generalist and lack task-specific fine-tuning, limiting their full potential. Additionally, while OpenAI’s closed-source o-series models support multimodal reasoning, DeepSeek-R1 does not. However, GRPO creates a path for open-source projects to convert any LLM or MLLM into a reasoning model \cite{openr1, openr1multimodal, openr1v}, enabling the generation of task-specific, multimodal, small reasoning models.

\noindent \textbf{Remote Sensing MLLMs.} Following the success of multimodal large language models (MLLM) on visual tasks, early models are exploring remote sensing capabilities. Hu et al. \cite{hu2307remote} introduce RSGPT for remote sensing imagery analysis, developing RSICap (2,585 human-annotated captions) and RSIEval benchmark. Kuckreja et al. \cite{kuckreja2024geochat} present GeoChat, an MLLM for interactive, spatially-aware satellite image analysis supporting multitask interactions through a 318k instruction dataset and LoRA-adapted fine-tuning \cite{Hu2022} for superior zero-shot generalization.

Zhang et al. \cite{zhang2024earthgpt} introduce EarthGPT, a multimodal LLM unifying remote sensing tasks through visual-enhanced perception and cross-modal comprehension, trained on the MMRS-1M dataset spanning optical, SAR, and infrared modalities. Bazi et al. \cite{bazi2024rs} present RS-LLaVA, an MLLM for captioning and VQA fine-tuned with LoRA on RS-instructions, outperforming previous models in multi-task efficiency. Muhtar et al. \cite{muhtar2024lhrs} develop LHRS-Bot, an MLLM leveraging volunteered geographic information and global RS images, supported by LHRS1-Align (1.15M image-caption pairs) and LHRS-Instruct datasets with GPT-4-generated reasoning tasks. These advances demonstrate AI-driven improvements in semantic reasoning and spatial comprehension for satellite imagery analysis.

While these models are generally designed for captioning or VQA and are guided by scene-specific prompts provided in the questions, they are not quantitatively tested for their ability to answer open-ended questions such as \textit{"Explain the image in detail."} Moreover, until now, no multimodal RL-powered reasoning model has been specifically designed for the RS field.

\section{Dataset}
\label{sec:dataset}

Constructing a high-quality dataset was essential for training and evaluating the models. The dataset in this study consists of satellite imagery with annotations indicating whether a military installation is present or not. The data was curated from two primary sources: (a) images likely containing military installations (specifically, surface-to-air missile (SAM) sites) and (b) images of civilian residential areas as negative examples.

\begin{figure}
    \centering
    \includegraphics[width=\linewidth]{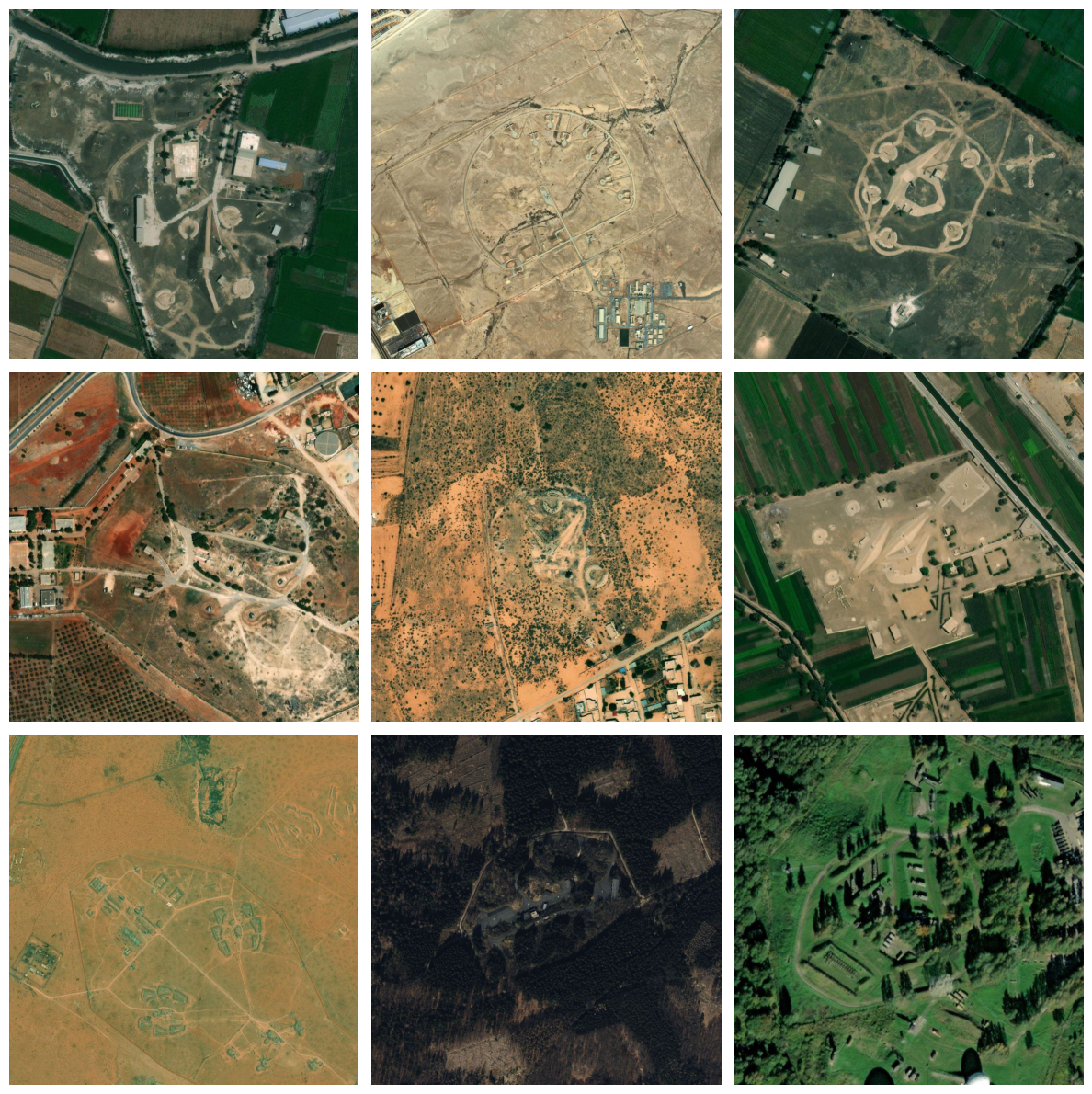}
    \caption{Some diverse examples of aerial imagery of SAM sites.}
    \label{fig:dataset}
\end{figure}

\noindent \textbf{Positive (Military) Examples:} A list of known or suspected military sites was obtained. In particular, an open-source KMZ file of global SAM sites was used \cite{samsites}, which contains coordinates of SAM launch batteries worldwide, contributed by satellite imagery enthusiasts. This yielded hundreds of candidate locations. Using an API for satellite imagery, high-resolution (1024x1024) overhead images at these coordinates were downloaded. By the help of a military aerial imagery expert, these images were analyzed to determine whether a military area was present in the current image or not. 318 images representing 129 distinct SAM sites have been collected. \textcolor{revision}{Although the number of SAM sites may appear relatively small for training a language model, they are distinctive examples drawn from 23 countries across 4 different continents, each reflecting diverse environmental conditions. Moreover,} these images vary in clarity--some show the classic circular layout of missile launch pads, while others are more subtle (e.g., small installations or partially obscured by terrain). \textcolor{revision}{As can be seen from the examples in Figure \ref{fig:dataset}, a model would require strong generalization ability to correctly identify each of these aerial images.}

\noindent \textbf{Negative (Civilian) Examples:} To train and test the model’s ability to avoid false alarms, a large set of images that do not contain military facilities was gathered. The coordinates of city centers globally were randomly sampled using a public “world cities” list \cite{worldcities}, and images around those areas were fetched. To avoid bias toward only urban scenes, slight random perturbations were added to the coordinates, yielding patches that might include outskirts, farmlands, or empty areas near cities. Initially, thousands of random images were collected to provide a broad variety of terrain and civilian infrastructure.

To caption these images, an open-source multimodal large language model (MLLM) was used as an annotator. Specifically, Qwen2-VL-72B (the largest model in the family) was prompted to generate a detailed caption of 4-6 sentences for each image. The prompt is always given as: \textit{"Explain the image in detail, with 4-6 sentences."} If the generated caption explicitly mentioned military-related words like “military”, “missile”, or “silo”, the image was tagged as containing a military installation and categorized as Category 0 (C0). The images where the model did not mention military terms were still kept as positive examples if they were labeled as military sites by a human expert. These 4 cases where the baseline model “fails” to identify the military nature of the site, which are classified as Category 1 (C1). Images sampled from negative examples are categorized as Category 2 (C2).

For the training set, \dataset-300-Train, 101 images from C0 and their captions from the Qwen2-VL model were selected as positive examples, while 200 randomly selected civilian images representative of residential, industrial, or natural scenes with no obvious military presence were added as negative examples (C2).

For the test set, \dataset-300-Test, 15 remaining images from C0 were selected to check verification performance, while all remaining 188 military images from C1--each from distinct locations not present in the training set--were included to measure the model’s performance on harder examples, on which the 72B model fails. Finally, 100 civilian images from C2 were added to assess whether the model was overfitting by incorrectly classifying non-military images as military.

\section{Method}
\label{sec:method}

\begin{figure*}[t!]
  \centering
  \includegraphics[width=\linewidth]{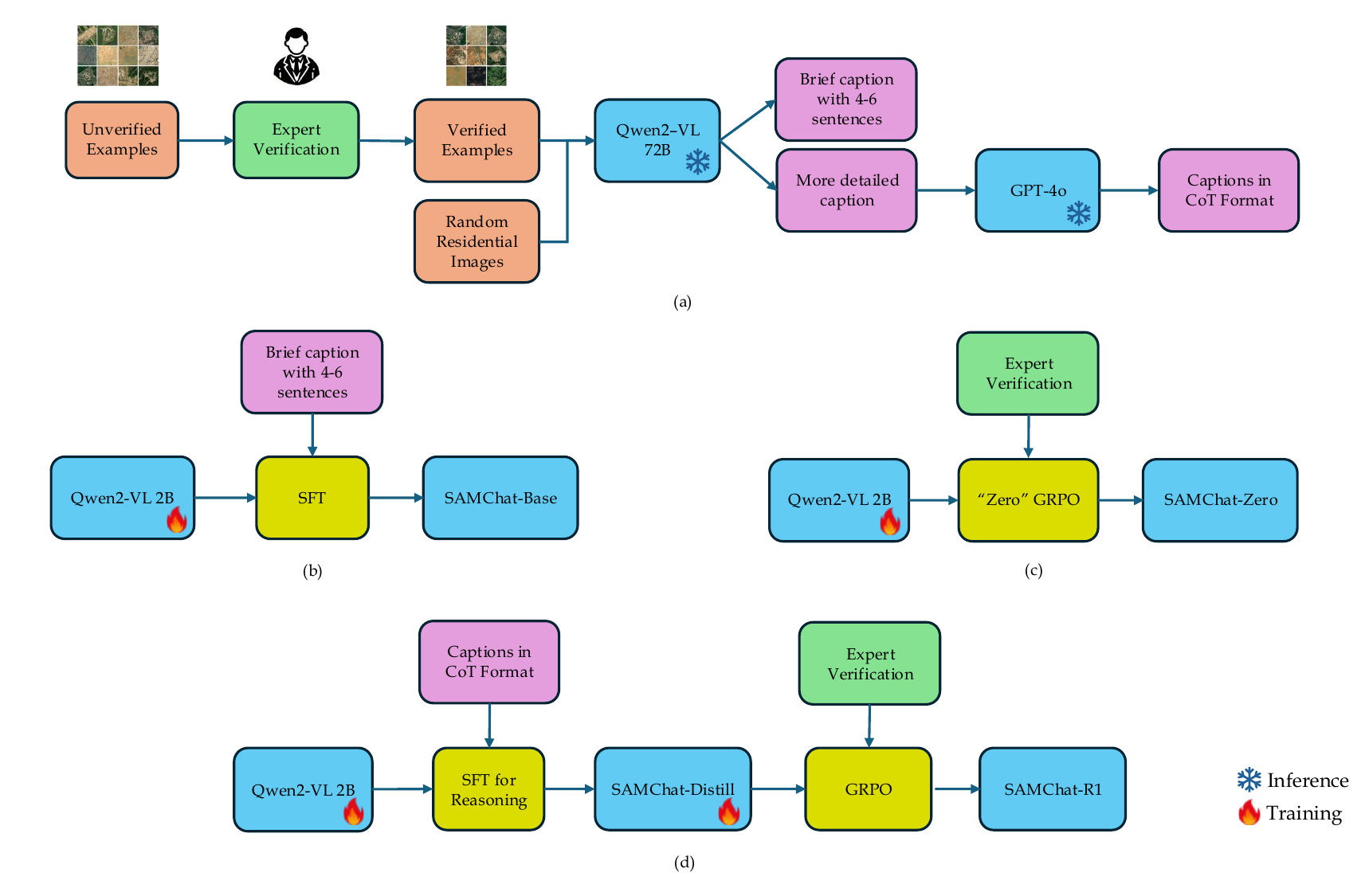}

   \caption{Training pipeline for the proposed \method family. \textcolor{revision}{(a) Satellite images containing expert-verified surface-to-air missile (SAM) sites and randomly sampled residential areas are processed by existing multimodal LLMs (MLLMs) to obtain both brief descriptions and chain-of-thought (CoT) captions. Starting from Qwen2-VL-2B, we train either (b) via supervised fine-tuning (SFT) on brief captions to obtain the base \method model, or (c) via zero-shot reinforcement learning using Group Relative Policy Optimization (GRPO) with only expert-verified yes/no labels (no captions) to obtain \method-Zero. (d) The final model, \method-R1, is produced by first applying SFT on CoT captions and then further aligning with GRPO.}}
   \label{fig:proposed}
   \vspace{-.5cm}
\end{figure*}

In this work, \method model family is proposed to harness the potential of MLLMs on RS images reasoning and captioning. This part explains the model arcitecture and training methodology for \method.  

\subsection{Model Architecture}

\method uses Qwen2-VL-2B \cite{wang2024qwen2} as the base model, with its pretrained weights and following its architecture:

\noindent \textbf{Visual Encoder:} Similar to Qwen2-VL-2B, the visual encoder in \method utilizes the Vision Transformer (ViT) architecture \cite{dosovitskiy2020image} with 675M parameters. During both training and inference, the encoder incorporates naive dynamic resolution support \cite{dehghani2023patch}, allowing it to process images of varying resolutions by dynamically converting them into a variable number of visual tokens. 

\noindent \textbf{Language Model:} Just like Qwen2-VL-2B, \method utilizes a large language model as its core component, initializing with pre-trained weights from Qwen-1.5B.

\noindent \textbf{Position-Aware Vision-Language Adapter:} Based on Qwen2-VL-2B, \method improves efficiency by incorporating a vision-language adapter that compresses visual features. Using a single-layer cross-attention module, it reduces the image feature sequence to a fixed length of 256 while preserving positional information with 2D absolute positional encodings. The compressed features are then fed into the large language model.

\subsection{Training}

\textbf{Supervised Fine Tuning (SFT):} Qwen2-VL-2B is fine-tuned on the training set of 300 image-caption pairs to generate the \method-Base model, where the captions have been generated via the larger version of the same model, Qwen2-VL-72B. This is a standard next-token prediction training where the caption is learned to be output by the model given the image. The objective is that after fine-tuning, given a new image, military-related keywords will be mentioned by the model's generated caption if appropriate. This approach turns the classification into a caption-generation problem, but it is later evaluated in binary terms (military or not) by keyword search in the outputs.

\noindent \textbf{SFT with Chain-of-Thought (CoT) Reasoning Captions:} It is hypothesized that training the model to produce this multi-step reasoning will help the model learn the explanation behind an image being military or not. In other words, instead of just stating “This is a military installation”, the model should first identify key features (e.g., “There are circular launch pads, a radar system, and a restricted perimeter”) and then conclude that the image likely represents a military site. This structured reasoning process helps the model learn the underlying logic behind why an image should be classified as military or not, making its predictions more reliable and interpretable. Prior work suggests that Chain of Thought (CoT) prompting can improve model reasoning and robustness for much larger language models \cite{wei2022chain}. On the other hand, for smaller models, persistently getting a zero-shot reasoning output just by prompting is not feasible. Instead, SFT with captions following CoT format may convert a small language model to a reasoning model.

For this purpose, this time, instead of 4-6 sentences, longer explanations are requested from the Qwen2-VL-72B model with the prompt \textit{"Explain this image in detail, as long as possible."} with 32,768 maximum token length. However, when this MLLM is prompted to convert these long paragraphs into a CoT format, the imitation of the thought processes were not diverse enough to train a small model. Therefore, these long outputs have been fed into GPT-4o, achieving more creative thought process outputs. These outputs were then used as captions for the images, and SFT is done as in the previous approach. With this training, \method-Distill model has been obtained.

\noindent \textbf{Reinforcement Learning with GRPO:} In addition to supervised fine-tuning, a lightweight reinforcement learning phase is experimented with to further align the model's outputs with the classification goal. Group Relative Policy Optimization (GRPO) is applied, which is a form of PPO that allows a group of generated answers to be used to compute relative rewards. \textcolor{revision}{The RL-based reward approach has proven highly beneficial in language models for enhancing reasoning capabilities, particularly in mathematical tasks, where there is only one correct answer and the reward is 1 if the answer is correct, and 0 otherwise. Our task can be naturally formulated in a similar binary reward setting. Hence,} our reward design is straightforward: the mentioning of military-related keywords is encouraged when appropriate, and discouraged on civilian images. Concretely, for a given image, multiple captions are generated from the model (with some randomness). If the image is a known positive (Category 0 in training), a higher reward is given to any caption that includes a military-related keyword. If the image is negative, higher reward is given to captions that remain free of such terms. This deterministic keyword-based reward is simple but encodes the classification objective. In addition to the keyword-based reward, the format is also rewarded. If the answer of the model follows the format \texttt{<reasoning> </reasoning> <answer> </answer>}, then the reward is given regardless of what is written within the reasoning process and the final answer. The policy is then updated by GRPO to increase the probability of high-reward generations. GRPO was selected over vanilla PPO because of its sample-efficient improvements on reasoning tasks reported by DeepSeek, meaning fewer RL steps and less risk of the model collapsing or forgetting its base knowledge can be achieved.

GRPO can be used as a standalone method for training, starting from the pretrained model without SFT and using RL with the defined reward function. This case tests whether the model can learn from a \textit{"zero start"}, per DeepSeek, via reward signals alone, as in the \method-Zero approach. Although this approach might be less stable given the small dataset, it is worth exploring to understand the effect of RL on small language models for remote sensing imagery. However, the primary focus should be on the combined method of SFT with reasoning and GRPO.

\noindent \textbf{SFT with CoT Reasoning Captions + GRPO:} A sequence of SFT + RL is explored, inspired by DeepSeek-R1's pipeline. First, SFT with CoT captions is performed to give the model reasoning ability. Then GRPO-based RL is run on the fine-tuned model for a small number of iterations to get the final model, \method-R1. The benefits of both are expected to be gained by the final model: the general descriptive and reasoning power from SFT and the sharpened focus from RL on the "military" cue.

After the training process, following the naming convention from the DeepSeek-R1 paper, the proposed \method model family is as follows:

\begin{itemize}
    \item \method-Base: Standard model, trained with SFT,
    \item \method-Zero: Reasoning model, trained from a zero shot start using GRPO,
    \item \method-Distill: CoT reasoning model, trained with SFT using CoT reasoning captions,
    \item \method-R1: CoT reasoning model, trained with SFT using CoT reasoning captions and GRPO.
\end{itemize}
\section{Experiments}
\label{sec:exp}

% \begin{figure}[htbp]
%   %\centering
%   \includegraphics[width=1\linewidth]{figures/qualitative_results.pdf}

%    \caption{Qualitative examples of the proposed model, \method-R1. While chat bubbles with dashed lines show the thinking process, those with the solid lines are the final answer. The one on the bottom right is a failure case, meanwhile the other three are the successful examples.}
%    \label{fig:qualitative}
% \end{figure}

\begin{revisionfigure*}[htbp]
\centering
\vspace{-.6cm}

\begin{tabular}{l}
\toprule

\scalebox{0.7}{\begin{minipage}[htbp]{1.45\linewidth}

\begin{wrapfigure}{l}{0.3\textwidth}
\includegraphics[width=0.3\textwidth]{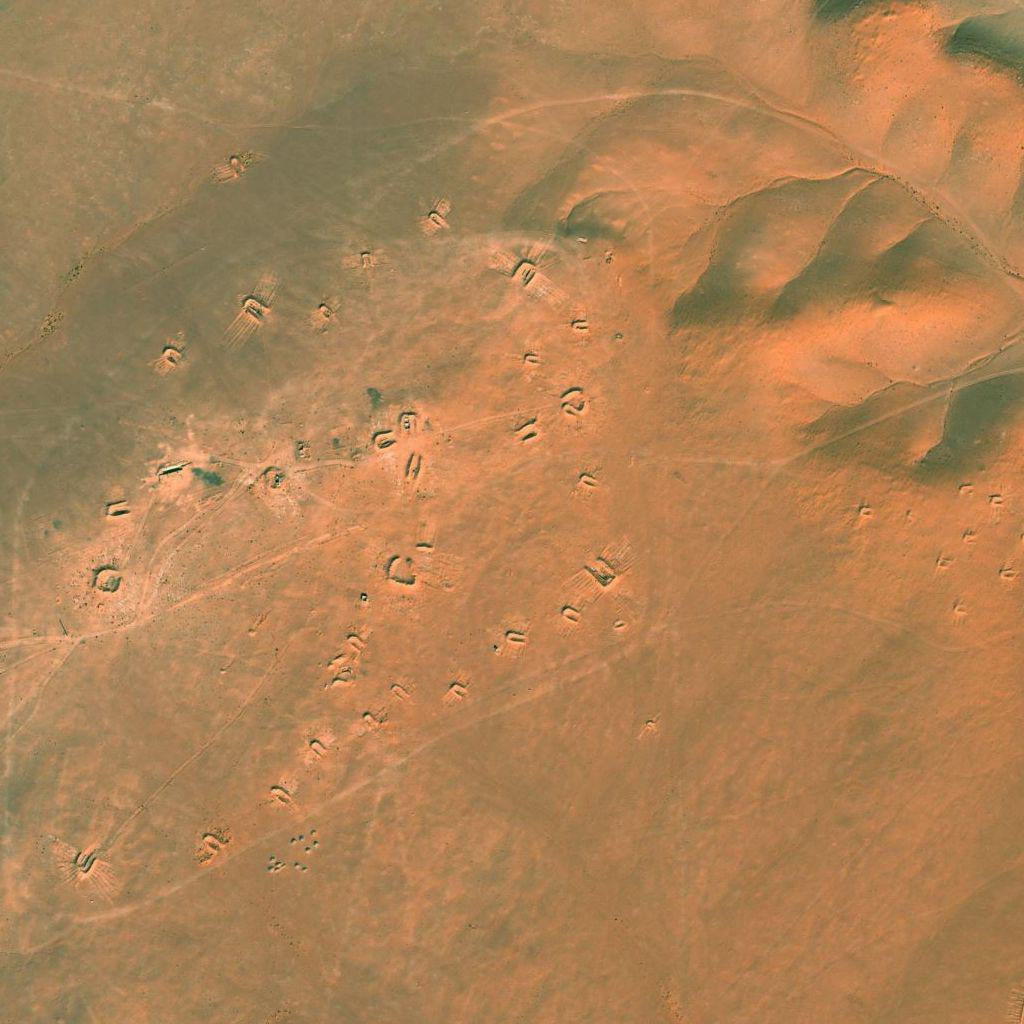}
\end{wrapfigure}

\underline{\textbf{\method (Base)} -- No CoT reasoning}

\textbf{Answer:} This image is an aerial or satellite view of a desert landscape, characterized by its arid and barren terrain. The area is dominated by a series of linear and circular patterns, which appear to be the result of human activity. These markings could be the remains of ancient irrigation systems, such as ditches or canals, used for agriculture in the past. The varying sizes and orientations of the patterns suggest a complex and organized system, likely designed to manage water resources efficiently. The surrounding landscape is mostly flat and sandy, with some gentle hills or ridges visible in the background. The overall color palette is dominated by shades of brown and beige, indicating a dry, possibly semi-arid environment.

\underline{\textbf{\method (+ CoT)}}

\textbf{Reasoning:} Alright, let me take a closer look at this image. The first thing that stands out is the reddish-brown color, which suggests a desert or arid environment. The terrain appears relatively flat with some gentle undulations, but the overall landscape looks barren and dry. There are no visible bodies of water, roads, or large structures, reinforcing the idea that this is a remote and sparsely populated area. Now, let me focus on the circular and rectangular formations. These structures are scattered across the landscape, and their arrangement suggests a deliberate pattern rather than random erosion. The circular formations are more concentrated in certain areas, while the rectangular ones are more evenly distributed. This could indicate human activity, possibly excavation or construction. Oh, I see something interesting—there are several lines connecting these structures. These lines seem to follow a specific pattern, which makes me think they might be roads or paths. They are not straight but rather follow the natural contours of the land, which suggests that the area is not completely flat. The presence of these roads indicates that people have accessed this location at some point, possibly for exploration, excavation, or construction. Wait, what about the surrounding environment? The terrain is mostly barren, with very little vegetation, which makes sense for a desert or semi-arid region. There are a few small structures scattered around, but they are not the main focus of the image. The overall color and texture suggest a dry, arid climate with minimal human activity except for the circular and rectangular formations. So, putting it all together, this image likely depicts a remote desert region with circular and rectangular formations, possibly remnants of human activity such as excavation or construction. The presence of roads suggests that the area has been accessed by people, but the lack of visible infrastructure indicates that it is not a major settlement. 

\textbf{Answer:} This image shows a remote desert region with a reddish-brown, arid landscape. The circular and rectangular formations are scattered across the terrain, with a pattern that suggests human activity, possibly excavation or construction. Several lines connect these structures, indicating roads or paths that follow the natural contours of the land. The surrounding environment is mostly barren, with very little vegetation, reinforcing the idea of a desert or semi-arid climate. The overall lack of visible infrastructure suggests that this area has not been heavily developed, with only sporadic human presence. 

\underline{\textbf{\method-R1 (+ GRPO)} }

\textbf{Reasoning:} Alright, let me take a closer look at this image. The first thing that stands out is the reddish-brown color, which suggests a desert or arid environment. The terrain appears relatively flat with some gentle undulations, though there are no visible hills or mountains. Now, let me focus on the structures. Oh, I see numerous circular and rectangular formations scattered across the landscape. These shapes are quite distinct and seem to be arranged in a somewhat organized manner. Some of them have a more symmetrical appearance, while others are more irregular. Wait, what could these be? They could be remnants of ancient structures, possibly ruins or defensive positions. But wait—some of them look like they might be man-made, possibly remnants of \textbf{military} installations or industrial sites. The circular formations might indicate bunkers or defensive positions, while the rectangular ones could be buildings or storage units. Now, let me think about the roads. I notice a network of lines crisscrossing the landscape. These roads are not straight but follow the natural contours of the land, which makes sense in a desert environment. They seem to be well-trodden, meaning they are frequently used. But wait—there are also some small, isolated structures scattered around, possibly remnants of past settlements or abandoned buildings. One more thing—there are some lighter patches scattered throughout the image. These could be water sources, possibly underground or in small, isolated areas. Their presence would explain the overall dryness of the region. So, putting all of this together, this image likely depicts a desert landscape with remnants of past human activity, possibly \textbf{military} or industrial in nature. The circular and rectangular formations suggest defensive positions or storage facilities, while the roads indicate frequent use and accessibility. The presence of small structures and water sources further supports the idea of human activity in this otherwise barren region. 

\textbf{Answer:} This image shows a desert landscape with a mix of natural and human-made features. The reddish-brown terrain suggests an arid environment, and the scattered circular and rectangular formations indicate remnants of past human activity, possibly \textbf{military} or industrial. The network of roads follows the natural contours, suggesting frequent use and accessibility. Small isolated structures and water sources further support the idea of human habitation in this otherwise barren region. The circular formations could be bunkers or defensive positions, while the rectangular ones might be buildings or storage units. Overall, this image presents a fascinating contrast between the harsh natural environment and the remnants of human activity. 

\end{minipage}}\\
\midrule

\scalebox{0.7}{\begin{minipage}[htbp]{1.45\linewidth}

\begin{wrapfigure}{l}{0.3\textwidth}
\includegraphics[width=0.3\textwidth]{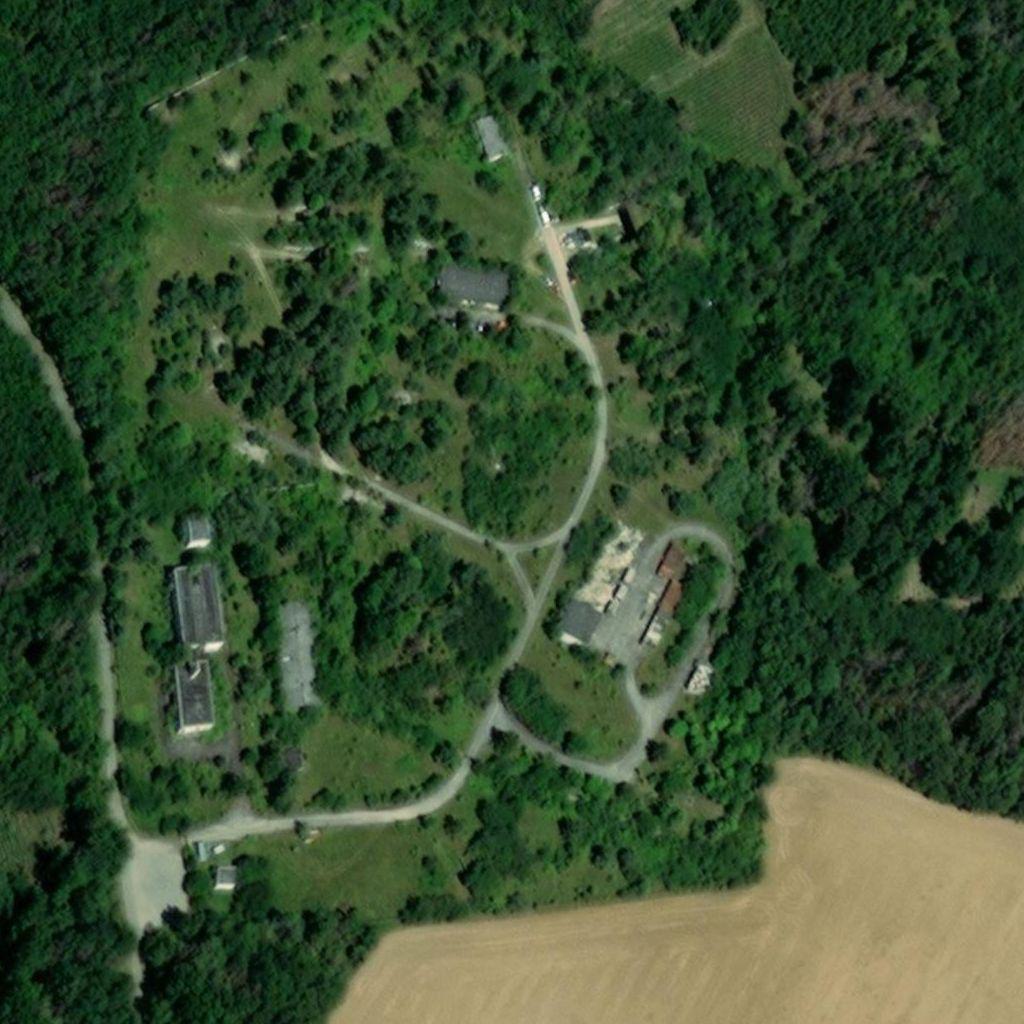}
\end{wrapfigure}

\underline{\textbf{\method (Base)} -- No CoT reasoning}

\textbf{Answer:} This image is an aerial view of a rural area surrounded by dense forest. The layout features a series of roads that intersect and wind through the landscape, connecting various buildings and structures. There are several large buildings, possibly homes or farmhouses, scattered throughout the area, with some buildings having dark roofs and others with lighter roofs. The area is interspersed with patches of open land and fields, indicating agricultural use. The overall scene suggests a remote, possibly agricultural or residential, setting with a strong emphasis on natural surroundings.

\underline{\textbf{\method (+ CoT)}}

\textbf{Reasoning:} Alright, let me take a closer look at this image. The first thing that stands out is the mix of natural and man-made elements. I can see a dense forest covering most of the area, with trees of various sizes and shades of green. This suggests a thriving ecosystem, possibly with different types of vegetation. Now, let me focus on the man-made structures. Oh, I notice several buildings scattered throughout the area. Some of them are rectangular, while others have more complex shapes. The buildings are not densely packed, which makes me think this is a rural or semi-rural setting. Some of the buildings are clustered together, while others are more isolated. Wait, I also see several roads and paths crisscrossing the area. The roads are not straight; they curve and branch off, which suggests a relatively small and unorganized settlement. Some of these roads appear to be dirt tracks, reinforcing the idea that this is a remote or less developed area. Now, let me think about the agricultural land. There is a noticeable patch of brownish land at the bottom of the image. This could be an agricultural field, possibly used for crops or livestock. The presence of trees around it suggests that the land might be used for forestry or grazing. Finally, I notice some open spaces within the forest and between the buildings. These could be parking areas, storage areas, or even small recreational areas. The overall layout suggests a rural or semi-rural community with a mix of residential, agricultural, and possibly industrial or research-related buildings. 

\textbf{Answer:} This image shows a rural or semi-rural area with a dense forest covering most of the landscape. Several buildings of different shapes and sizes are scattered throughout, with some clustered together and others more isolated. The roads are not strictly straight but curve and branch, indicating a small and unorganized settlement. A patch of brownish land at the bottom suggests an agricultural field, possibly used for crops or livestock. The presence of trees and open spaces suggests a mix of natural and human-made elements, with a possible focus on agriculture, forestry, or research. Overall, this appears to be a remote and sparsely populated area with a combination of residential, agricultural, and possibly industrial or research-related structures. 

\underline{\textbf{\method-R1 (+ GRPO)} }

\textbf{Reasoning:} Alright, let me take a closer look at this image. The first thing that stands out is the mix of natural and man-made elements. I can see a dense forest covering most of the area, with trees of various sizes and shades of green. This suggests a thriving ecosystem, possibly with different types of vegetation. Now, let me focus on the man-made structures. Oh, I notice several buildings scattered throughout the area. Some of them are rectangular, while others have more irregular shapes. The buildings are not densely packed, which makes me think this is a rural or semi-rural location. Some of the buildings are connected by winding roads, which indicate a network of paths for access. The roads are not perfectly straight; they curve and branch off, which suggests a less structured, more organic layout. Wait, I also see a large open area at the bottom of the image. It looks like a field or a meadow, with a distinct boundary. This could be used for agriculture or grazing. The presence of a clear boundary suggests it might be a managed land, possibly a farm or a hunting ground. Now, let me think about the overall purpose of this place. Given the mix of buildings, roads, and natural surroundings, it could be a small rural settlement, possibly a farm, a research facility, or even a \textbf{military} installation. The presence of multiple buildings and the winding roads suggests a place where people live or work, but not in a highly structured urban setting.

\textbf{Answer:} This image shows a rural or semi-rural area with a dense forest covering most of the landscape. Several buildings are scattered throughout, connected by winding roads that curve and branch off, indicating a less structured, organic layout. A large open area at the bottom suggests a field or meadow, possibly used for agriculture or grazing. The presence of multiple buildings and the winding roads suggests a place where people live or work, but not in a highly structured urban setting. Overall, this appears to be a small rural settlement or facility, possibly a farm, a research facility, or even a \textbf{military} installation. 

\end{minipage}}
\\

\bottomrule
\end{tabular}

\caption{Qualitative comparison of base \method, \method with CoT fine-tuning and \method-R1 responses for two examples of \dataset-300-Test dataset. Reasoning steps are shown above the final concise answer.}
\label{fig:qualitative}
\end{revisionfigure*}

\subsection{Implementation Details}

The training pipeline is implemented using PyTorch and the HuggingFace Transformers library, leveraging pretrained weights for Qwen2-VL models. The supervised fine-tuning parts were done on HPC clusters with 4 $\times$ NVIDIA H100 GPUs (SXM5 64GB), which was sufficient for full parameter fine-tuning of the 2B model. Adam Optimizer is used with an initial learning rate of $1 \times 10^{-5}$ and a batch size of 16, for 3 epochs for standard fine-tuning and 6 epochs for CoT fine-tuning. 

For the RL stage, the TRL (Transformer Reinforcement Learning) framework is used to implement GRPO updates in a batch fashion. The training is done on HPC clusters with 8 $\times$ NVIDIA A100 GPUs (80GB), Adam optimizer with initial learning rate of $1 \times 10^{-6}$ and a batch size of 8, 4 samples per image for GRPO. Training is performed for five epochs for GRPO after SFT, which amounts to 6,000 training episodes, and six epochs for "zero" start GRPO, which amounts to 7,200 training episodes. 

\subsection{Evaluation Metrics}

Precision, Recall, and F1-score are primarily reported for the number of images that have military-related keywords ("military, missile, silo") in the output captions, since this is essentially a detection problem. Performance is also broken down by the test categories defined:

\begin{itemize}
    \item Easy positives (Category 0): number of correctly identified out of 15 (should be high).
    \item Hard positives (Category 1): recall on these 188 difficult cases (the main metric to improve).
    \item Negatives (Category 2): number of false positives out of 100 (ideally zero is desired).
\end{itemize}

These category-specific results help in understanding where fine-tuning helps most. An ideal model would score high on both Category 0 and 1 while maintaining near 0 false alarms on Category 2. Hence, Recall will be computed over all positives (Categories 0+1, total 203 images), while Precision metric will be calculated over all negative examples (Category 2). For reasoning models, only the sentences between \texttt{<answer> </answer>} are considered; the reasoning part is omitted.

\begin{table}[htbp]
    \centering
    \caption{Detection performance (\%) of available methods in the literature. "Prop." is abbreviation for proprietary, shows closed source models while the detection performances are measured via API. RS shows model is adapted to remote sensing problems with domain specific data. "Sonnet Th."  denotes Claude Sonnet 3.7 Extended Thinking.}
    %\scalebox{0.9}{
    \begin{tabular}{l|c|c|c|c|c}
    \toprule
        Model & Param & RS & Recall & Precision & F1 \\    
    \midrule
        Qwen2-VL \cite{wang2024qwen2} & 76B & $\times$ & 7.4 & 99.0 & 13.8 \\
        Qwen2-VL \cite{wang2024qwen2} & 2B & $\times$ & 19.2 & 99.0 & 32.2 \\
        GPT-4o \cite{hurst2024gpt} & Prop. & $\times$ & 32.0 & \textbf{100.0} & 48.5 \\
        GPT-4.5 \cite{gpt45} & Prop. & $\times$ & 27.6 & \textbf{100.0} & 43.2 \\
        Sonnet 3.5 \cite{sonnet35} & Prop. & $\times$ & 33.0 & 99.0 & 49.5 \\
        Sonnet 3.7 \cite{sonnet37} & Prop. & $\times$ & 16.3 & 99.0 & 27.9 \\
        RS-LLaVA \cite{bazi2024rs} & 7B & \checkmark & 0.0 & \textbf{100.0} & 0.0 \\
        GeoChat \cite{kuckreja2024geochat} & 7B & \checkmark & 0.0 & \textbf{100.0} & 0.0 \\
        LHRS-Bot \cite{muhtar2024lhrs} & 7B & \checkmark & 40.4 & \textbf{100.0} & 57.5 \\
        \method-Base & 2B & \checkmark & \textbf{47.8} & 99.0 & \textbf{64.5} \\
        \midrule
        Sonnet Th. \cite{sonnetthinking} & Prop. & $\times$ & 19.7 & \textbf{100.0} & 32.9 \\
        o1 \cite{o1preview} & Prop. & $\times$ & 13.3 & \textbf{100.0} & 23.5 \\
        \method-R1 & 2B & \checkmark & \textbf{80.8} & 98.0 & \textbf{88.6} \\
        \bottomrule
    \end{tabular}%}
    \label{tab:results}
\end{table}

\subsection{Experimental Results}

The fine-tuned small models indeed outperform the non-fine-tuned large models on the task of military installation classification. Table \ref{tab:results} summarizes the performance metrics, and some qualitative examples of the proposed reasoning model, \method-R1 are provided in Figure \ref{fig:qualitative}.

Most important key observations of Table \ref{tab:results} are as follows:

\begin{itemize}
    \item The SFT 2B model (\method-Base) achieved a recall of about 48\%, more than 6$\times$ the recall of the 72B model (7.4\%) on this dataset, while keeping precision at 99\%. In other words, it detected many more of the secluded military sites than the large model did, and did not mislabel any more civilian images as military. This confirms the hypothesis of the study that targeted fine-tuning can unlock a small model’s capability in ways a large general model does not do out-of-the-box. It is noteworthy that even the zero-shot 2B model had higher recall (19.2\%) than the 72B -- this indicates some intrinsic differences in how the smaller model operated (perhaps it was more “eager” to guess military in its pretraining distribution, or simply randomness). Fine-tuning further doubled that recall.
    \item The most popular proprietary general-purpose MLLMs, GPT-4o and Claude Sonnet 3.5,  got similar scores, around 33\% recall, 99\% precision and 49\% F1 score. Interestingly, both of their newer versions, GPT-4.5 and Sonnet 3.7, respectively, performed even worse than the older ones. Considering the proposed base model, \method-Base, the difference is clear: a 2B open model caught roughly 97 out of 203 military sites, whereas GPT-4 and Claude caught maybe 68 or fewer. This result underscores the value of domain-specific training even relative to the most advanced general AI models of today.
    \item Probably due to being focused on residential areas, some of the existing domain-specific language models such as GeoChat and RS-LLaVA do not caption any \dataset-Test images with military-related keywords, which leads to 0 recall and F1 scores. Both of these algorithms can caption airports, which are also counted as residential areas, excluded from the military-related keywords on purpose. On the other hand, a more recent approach, LHRS-Bot gives more competitive results on the dataset. With its 40.4\% recall and 100\% precision score, this language model is the second best among the standard models, outperforming proprietary large models.
    \item Reasoning ability brings the base performance \method to another level. The proposed task-specific multimodal reasoning model, \method-R1, outperforms the \method-Base by almost 2$\times$, proprietary generalist Sonnet 3.7 Extended thinking by 4$\times$, and o1 by 6$\times$ with the same number of parameters as the base model. o1, which performs worse than the normal language models GPT-4o and GPT-4.5, is probably trained for math and coding-based tasks, overlooking the complexities in aerial images, even though it is defined as a generalist model. This shows that both CoT reasoning-based SFT and GRPO ideas are beneficial in the RS domain. Further ablation studies will verify this observation.
\end{itemize}

\noindent \textbf{Failure Cases.} While \method-R1 performs well, failure cases persist, particularly false negatives from camouflage or obstructions and false positives from similar civilian structures. Dense vegetation, low resolution, or concealed details can obscure missile sites, while industrial facilities or circular fields may resemble military targets. These errors suggest \method-R1 relies heavily on geometric cues, which are sometimes insufficient. Future improvements could refine training data with more challenging examples or integrate human-in-the-loop validation for better robustness.

\section{Ablation Studies}
\subsection{Effect of Training Methods}

The effect of methods mentioned in Section \ref{sec:method} are analyzed in detail in Table \ref{tab:method_ablation}. 

\begin{table}[htbp]
    \centering
    \caption{Number of the positive examples for each category in each proposed model. Numbers in the parentheses show the total number of images in that category. For C0 and C1, the positives are true, for C2 they are false.}
    %\scalebox{0.9}{
    \begin{tabular}{l|c|c|c}
        \toprule
        Method & C0 (15) & C1 (188) & C2 (100) \\
        \midrule
        Zero-Shot & 5 & 34 & 1 \\
        SFT & 11 & 86 & 1 \\
        SFT for CoT Reasoning & 13 & 135 & 4 \\
        Zero GRPO  & 14 & 144 & 4 \\
        SFT for CoT Reasoning + GRPO & 15 & 149 & 2 \\
        \bottomrule
    \end{tabular}%}
    \label{tab:method_ablation}
    %\vspace{-.5cm}
\end{table}

\noindent \textbf{Impact of Chain-of-Thought:} The ablation comparing SFT vs SFT for CoT Reasoning  showed that the model with CoT reasoning in its training captions achieved more than 1.5$\times$ military-related answers, while the precision drops by 3\%, due to increasing false positives in C2. The intermediate reasoning in the CoT training likely taught the model what clues to look for. This aligns with observations that CoT can make models better at justifying and thereby correctly executing a task. Thus, incorporating reasoning-focused data is beneficial for fine-tuning multimodal models in this context.

\noindent \textbf{Impact of "Zero" GRPO:} While it is expected to be an unstable training due to both language model and training set being small, "zero" start GRPO training handled the problem well. Without any finetuning captions, the model outperforms the one which is supervised fine-tuned with CoT reasoning data, while the precision stays the same. This result shows that even without any training captions generated by either humans or existing language models, a small language model can perform competitively, especially on the RS images.

\noindent \textbf{Impact of CoT SFT + GRPO:} Applying GRPO on a supervised fine-tuned model with CoT reasoning captions can further increase the overall performance. According to the results, for the first time, a small language model can caption all military imagery in Category 0, while both recall and precision performance increase, in Categories 1 and 2, respectively. \textcolor{revision}{In numerical terms, the test data show 2 and 14 additional true positives in C0 and C1 (approximately +8\% in recall), along with 2 fewer false alarms in C2 (+2\% precision), achieved by applying GRPO-based RL over CoT SFT on the training set.} It can be deduced that, for optimal performance on this problem, RL gives the best result when used as a training on top of supervised fine-tuning, rather alone, for such tasks. 

\textcolor{revision}{\subsection{Applicability of the Method to Different Models}}

\textcolor{revision}{In this section, we analyze the applicability of the proposed method across different base models. These tests include models with the same architecture but larger parameter sizes, models with different architectures of equal or smaller parameter sizes, and a model already pre-trained on general remote sensing tasks. The training results of these base models, obtained from concise SFT training alone and from combined CoT SFT + GRPO training, are reported separately in Table \ref{tab:different_models}.}

\begin{revisiontable}[htbp]
    \centering
    \caption{Detection performance (\%) of the proposed method on different models. R1 indicates whether the model is trained with concise SFT ($\times$) or with CoT SFT + GRPO (\checkmark). Both GeoChat and LLaVA-GeoChat are trained using the GeoChat training code: GeoChat is initialized with GeoChat parameters, whereas LLaVA-GeoChat is initialized with LLaVA-1.5 parameters.}
    %\scalebox{0.9}{
    \begin{revisiontabular}{l|c|c|c|c|c}
    \toprule
        Base Model & R1 & Recall & Precision & F1 \\    
    \midrule
        Qwen2-VL-7B \cite{wang2024qwen2}  & $\times$ & \textbf{57.1} & 97.0 & \textbf{71.9} \\
        Intern-VL2-2B \cite{chen2024internvl} & $\times$ & 50.7 & 97.0 & 66.6 \\
        GeoChat \cite{kuckreja2024geochat} & $\times$ & 20.7 & 91.0 & 33.7 \\
        LLaVA-GeoChat \cite{liu2023llava} & $\times$ & 40.4 & 86.0 & 55.0 \\
        Qwen2-VL-2B (\method-Base) & $\times$ & 47.8 & \textbf{99.0} & 64.5 \\
        \midrule
        Qwen2-VL-7B \cite{wang2024qwen2}  & \checkmark & 77.8 & 96.0 & 86.0 \\
        Intern-VL2-2B \cite{chen2024internvl} & \checkmark & 80.3 & 96.0 & 87.5 \\
        Qwen2-VL-2B (\method-R1) & \checkmark & \textbf{80.8} & \textbf{98.0} & \textbf{88.6} \\
        \bottomrule
    \end{revisiontabular}%}
    \label{tab:different_models}
\end{revisiontable}

\textcolor{revision}{\noindent \textbf{Larger model.} To investigate the impact of scaling on Chain-of-Thought reasoning in the remote sensing domain, we trained a larger model from the same family and architecture, Qwen2-VL-7B, using the methodology proposed in this work. The results in Table \ref{tab:different_models} show that the 7B model achieves higher scores than its 2B counterpart when trained with concise fine-tuning data. However, while the combination of CoT SFT and GRPO training still improves performance—confirming that our method scales to larger models--the overall performance saturates slightly below that of the 2B model. This suggests that \method reaches its peak effectiveness at the 2B scale, and given the current limitations in data and the number of scenes, the 7B model does not provide additional gains for these military imagery tasks.}

\textcolor{revision}{\noindent \textbf{Same size, different model family.} To assess the reproducibility of our proposed method across different language model architectures, we applied the same training strategy to InternVL2-2B \cite{chen2024internvl}. The results in Table \ref{tab:different_models} show that the trained InternVL2-2B performs on par with our Qwen2-VL--based \method--slightly better in the concise model setting, and slightly worse in the reasoning model setting. These findings demonstrate that our approach can be effectively applied to other similarly sized open-source MSLMs.}

\textcolor{revision}{\noindent \textbf{Smaller model.} To evaluate the applicability of our method to smaller models, we examined existing 1B-scale models. Since Qwen2-VL does not provide a 1B variant, we proceeded with InternVL2-1B. However, preliminary supervised fine-tuning experiments were unsuccessful. InternVL2-1B employs Qwen2-0.5B as its LLM backbone, but the English corpus of the 0.5B model appears insufficient for handling complex imagery, such as aerial views of military installations. Moreover, as Qwen2 is a multilingual model with Chinese as its primary language, it often reverts to Chinese in challenging scenarios, even after fine-tuning with English questions and answers. Consequently, the results for InternVL2-1B cannot be included in Table \ref{tab:different_models}.}

\textcolor{revision}{\noindent \textbf{Fine-tuning existing RS model.} Instead of fine-tuning a general-purpose MSLM, training an existing pre-trained RS model provides a more intuitive approach for achieving better results. To this end, we employed the GeoChat model \cite{kuckreja2024geochat}. GeoChat was supervised fine-tuned with both concise and CoT captions, using GeoChat-7B and LLaVA 1.5-7B \cite{liu2023llava} as the pretrained backbones. According to the results in Table \ref{tab:different_models}, while GeoChat demonstrates a clear improvement over its zero-shot counterpart reported in Table \ref{tab:results}, it still lags considerably behind other concise models. The performance of reasoning GeoChat, on the other hand, could not be evaluated. Since both LLaVA and GeoChat are pretrained with a maximum sequence length of 2048 tokens, the model output terminates before completing its reasoning process. In contrast, reasoning captions in \dataset typically require a maximum sequence length of 4096--8192 tokens, which Qwen2-VL and InternVL2 can support, but LLaVA-GeoChat cannot. Therefore, reasoning results for GeoChat are omitted.}

\subsection{Effect of Prompting Methods}

\textcolor{revision}{Throughout the data creation and training process, we adopted open-ended prompting because it is more challenging, less biased by the question format, and provides a fairer assessment of a model’s true captioning and reasoning ability. One may wonder, however, how existing open-source models would perform if easier prompting strategies -- commonly used in the remote sensing LLM literature -- were applied instead. To investigate this, Table \ref{tab:prompting_ablation} presents the number of military images correctly identified by the open-source multimodal language model (MLLM) Qwen2-VL 72B under different prompting methods in a zero-shot setting. The results clearly demonstrate that prompting the model with the open-ended question \textit{“Explain the image.”} makes the task significantly more challenging in complex scenes compared to leading prompts such as \textit{“Is this a military area?”} or \textit{“Choose the purpose of the area: A. Military…”} In the latter cases, the prompt itself provides a strong hint, effectively turning the classification into a verification task. Therefore, we rely on the open-ended formulation in our main evaluation, as it better reflects the actual reasoning capability of MLLMs without inflating performance through prompt-specific shortcuts.}

\begin{table}[htbp]
    \centering
    \caption{The number of the images Qwen2-VL 72B correctly identifies with different prompting methods. Number in the parentheses show the total number of military images.}
    \begin{tabular}{l|c}
        \toprule
        Prompting Method & \# Correct (318) \\
        \midrule
        Open Ended & 101 \\
        Yes/No Question & 245 \\
         Multiple Choice Question & 219 \\
         \bottomrule
    \end{tabular}
    \label{tab:prompting_ablation}
    \vspace{-.5cm}
\end{table}
\section{Conclusion}
\label{sec:conc}

In this paper, a small-scale multimodal language model, \method, has been presented, adapted for open-ended reasoning on remote sensing images with a focus on identifying military installations in secluded areas. The \dataset dataset was created, and chain-of-thought annotations were employed to demonstrate how smaller models can surpass far larger counterparts on domain-focused tasks by combining domain-specific supervised fine-tuning with reinforcement learning via GRPO. Through this approach, detection performance was improved, and interpretable, step-by-step rationales--essential for expert validation and real-time decision-making scenarios--were provided. High precision and robust recall were attained by \method in experimental evaluations, reinforcing its practicality for deployment in resource-constrained or security-sensitive environments. Despite the strong performance of \method-R1 in detecting military sites, challenges remain in handling camouflaged locations and visually similar civilian structures. Future work could address these limitations through multi-modal data integration and human-in-the-loop validation.

\section*{Acknowledgements}

The numerical calculations reported in this paper were partially performed using the
MareNostrum 5 pre-exascale supercomputing system and TÜBİTAK ULAKBİM, High Performance and Grid Computing Center (TRUBA resources). We gratefully acknowledge the Barcelona
Supercomputing Center (BSC) and the Scientific and Technological Research Council of Turkey
(TÜBİTAK) for providing access to these resources and supporting this research.

We also appreciate the support of Col. (Ret.) Tamer Karaşahin on annotation verifications. 

% Our acknowledgments are to be shared. $\quad \quad \quad \quad$

%\section*{Acknowledgments}
%This should be a simple paragraph before the References to thank those individuals and institutions who have supported your work on this article.

\bibliographystyle{IEEEtran}
\bibliography{main}

\begin{IEEEbiography}[{\includegraphics[width=1in,height=1.25in,clip,keepaspectratio]{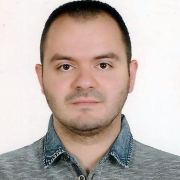}}]{Aybora Köksal}
is a PhD student at Electrical and Electronics Engineering Department of Middle East Technical University (METU). He has pursued his MSc studies at the same department with the focus on moving object detection and semi-supervised automated annotation methods. He is a researcher at METU Center for Image Analysis (OGAM). He is interested in computer vision, machine learning and image processing. He had received his two BSc degrees in the EEE and Mathematics departments of METU, with high honors.

His current research interests include multimodal large language models and their applications in remote sensing.
\end{IEEEbiography}

\begin{IEEEbiography}
[{\includegraphics[width=1in,height=1.25in,clip,keepaspectratio]{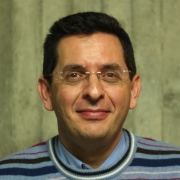}}]{A. Aydın Alatan}
received his B.S. degree from Middle East Technical University, Ankara Turkey in 1990, the M.S and DIC degrees from Imperial College of Science, Medicine and Technology, London, UK in 1992, and PhD degree from Bilkent University, Ankara Turkey in 1997, all in Electrical Engineering.

He was a post-doctoral research associate at Center for Image Processing Research at Rensselaer Polytechnic Institute between 1997 and 1998 and at New Jersey Center for Multimedia Research at New Jersey Institute of Technology between 1998 and 2000. In August 2000, he joined faculty of Electrical and Electronics Engineering Department at Middle East Technical University. He is the founding director of METU Center for Image Analysis (OGAM).

He is a senior member of IEEE and The Science Academy, Turkey (Bilim Akademisi).
\end{IEEEbiography}

\vfill

\end{document}